\documentclass[journal,10pt,oneside,twocolumn]{IEEEtran}

\usepackage{cite}

\usepackage{amsmath}
\usepackage{amsfonts}
\usepackage{booktabs}
\usepackage{multirow}
\usepackage{graphicx}

\usepackage{subfig}
\usepackage{xcolor}

\newcommand\hlbreakable[1]{\textcolor{black}{#1}}
\usepackage[switch,displaymath]{lineno}

\begin{document}
\title{Self-supervised Adaptive Weighting for Cooperative Perception in V2V Communications}

\author{Chenguang~Liu,  Jianjun~Chen, Yunfei~Chen, {\em{Senior~Member,~IEEE}},\\
Ryan~Payton, Michael~Riley, and~Shuang-Hua~Yang, {\em{Senior~Member,~IEEE}}

\thanks{This work was supported in part by Oracle Cloud credits and related resources provided by the Oracle for Research program, the National Natural Science Foundation of China (Grant No. 92067109, 61873119, 62211530106), and Shenzhen Science and Technology Program (Grant No. ZDSYS20210623092007023, GJHZ20210705141808024). (\it{Corresponding author: Shuang-Hua Yang.})}
\thanks{Chenguang Liu is with the School of Engineering, University of Warwick, Coventry, UK, CV4 7AL. {e-mail: chenguang.liu@warwick.ac.uk}}
\thanks{Jianjun Chen is with the Faculty of Engineering and Information Technology, University of Technology Sydney, Australia. {e-mail: jianjun.chen@student.uts.edu.au}}
\thanks{Yunfei Chen is with the Department of Engineering, University of Durham, Durham, UK, DH1 3LE. {e-mail: yunfei.chen@durham.ac.uk}}
\thanks{Ryan Payton and Michael Riley are with Oracle for Research. {email: ryan.payton@oracle.com; michael.riley@oracle.com}}
\thanks{Shuang-Hua Yang is with Shenzhen Key laboratory of Safety and Security for Next Generation of Industrial Internet, Southern University of Science and Technology, Shenzhen, China, and also with Department of Computer Science, University of Reading, UK. {e-mail: yangsh@sustech.edu.cn}}
}
\markboth{Journal of \LaTeX\ Class Files,~Vol.~14, No.~8, August~2015}%
{Liu \MakeLowercase{\textit{et al.}}: Self-supervised Adaptive Weighting for Cooperative Perception in V2V Communications}

\maketitle

\begin{abstract}
Perception of the driving environment is critical for collision avoidance and route planning to ensure driving safety. Cooperative perception has been widely studied as an effective approach to addressing the shortcomings of single-vehicle perception. However, the practical limitations of vehicle-to-vehicle (V2V) communications have not been adequately investigated. In particular, current cooperative fusion models rely on supervised models and do not address dynamic performance degradation caused by arbitrary channel impairments. In this paper, a self-supervised adaptive weighting model is proposed for intermediate fusion to mitigate the adverse effects of channel distortion. The performance of cooperative perception is investigated in different system settings. Rician fading and imperfect channel state information (CSI) are also considered. Numerical results demonstrate that the proposed adaptive weighting algorithm significantly outperforms the benchmarks without weighting. Visualization examples validate that the proposed weighting algorithm can flexibly adapt to various channel conditions. Moreover, the adaptive weighting algorithm demonstrates good generalization to untrained channels and test datasets from different domains.  

\end{abstract}


\IEEEpeerreviewmaketitle

\section{Introduction}
Cooperative perception enabled by vehicular communications facilitates the exchange of complementary perceptual information among multiple connected and autonomous vehicles (CAVs), leading to a more comprehensive and integrated perception for autonomous driving \cite{9963987}. By aggregating the information from multiple viewpoints, it provides an effective framework to alleviate the limitations of single-vehicle perception, to detect occluded or distantly located objects. Such collaboration schemes highly rely on the vehicular communications system to transmit the shared information from CAVs to the ego vehicle. In practice, the reliability of vehicle-to-vehicle (V2V) communications is constrained by limited bandwidth and channel distortion. Therefore, addressing the practical limitations imposed by the V2V communications system on collaborative perception is vital, which has garnered considerable interest from researchers. 

Based on the previous works in 3D detection backbones \cite{8578570, pointpillars, s18103337}, related works have been widely conducted to develop efficient collaborative fusion schemes, including sharing raw point clouds (\textit{i.e.} early fusion) \cite{8885377}, intermediate features (\textit{i.e.} intermediate fusion) \cite{10.1145/3318216.3363300, Wang2020V2VNetVC,9156848,where2comm,li2021learning,9812038, xu2022v2xvit,10077757,Qiao2023WACV} and detection results (\textit{i.e.} late fusion)\cite{9228884}. Most of these works focus on improving perception accuracy and reducing bandwidth utilization, assuming ideal exchange of information. To consider more realistic V2V communication, some studies examined the effects of communications delay \cite{xu2022v2xvit}, pose error \cite{Wang2020V2VNetVC}, and lossy communications \cite{10077757}. However, they have still assumed perfect communications channel without any channel impairments, except \cite{10184097}, which considered Rician fading and free-space path loss by incorporating the learning-based communication channel into cooperative perception. Moreover, the work in \cite{10184097} demonstrated that the intermediate features are more robust to channel impairments than raw point clouds and detection results. Also, the intermediate fusion can be further improved in the presence of channel impairments. 

In realistic V2V communications, the information shared among vehicles could potentially enhance the detection performance, but if it suffers from severe channel distortion, it may instead manifest as interference that undermines detection accuracy. Especially for intermediate features, it is more challenging to identify the corrupted features than the raw point clouds and detection outputs since latent features do not have a physical detection range. Therefore, it is vital to prevent the severely distorted information from taking part in the fusion or collaboration.

On the other hand, the channel distortion in realistic V2V communication may exhibit arbitrary and dynamic characteristics, leading to signal strength fluctuation \cite{5710954}. This non-stationarity of the V2V channel is mainly caused by the high velocity of the transmitter, receiver and moving scatterers in the V2V environment \cite{5307472}. This necessitates the cooperative system to adapt to various levels of channel impairment. In this regard, relying solely on the supervised training with the well-labeled dataset is inefficient when dealing with various communication environments, since labeling the received intermediate features is difficult due to their latent representation. Consequently, the cost of training and labeling would exponentially increase when scaling to a more complex communication environment or different levels of channel impairments. Supervised models might also encounter difficulties in generalizing effectively to unseen data that significantly differs from the training samples due to random distortion. Therefore, it is necessary to design a specialized loss function and training scheme to address these challenges.  


Although the previous works have explored various fusion schemes for cooperative perception, several challenges remain. For example, although different fusion schemes for cooperative perception with V2V communications were studied in \cite{10184097}, the significant performance degradation for intermediate fusion due to channel distortion has yet to be addressed. Additionally, the works mentioned above mainly use supervised learning with well-labeled datasets, but this could be impractical for dynamic and random channel distortions. Motivated by these, this work will focus on mitigating the adverse effects of channel distortion on the intermediate fusion in cooperative perception. The main contributions are summarized below:
\begin{enumerate}
    \item To address the severe performance degradation of intermediate fusion in \cite{10184097}, a self-supervised adaptive weighting model is proposed for the cooperative perception with intermediate fusion. A convolutional neural network (CNN)-based structure ending with Softmax is used for adaptive weighting. Self-supervised training with contrasting information is used to train this model without manual annotation. Different data domains, detection backbones, noise levels, and path loss factors are examined to validate its generalization and adaptive ability. 
    \item Unlike previous works that neglect realistic channel models, we evaluate the proposed adaptive weighting considering the WINNER II channel using orthogonal frequency-division multiplexing (OFDM). Different fusion methods and transmitted information are evaluated for different SNRs and path loss factors. A real-world dataset for cooperative perception is also considered.
    \item Numerical results show that the proposed adaptive weighting algorithm performs better than the benchmarks without weighting. It is also validated that the proposed adaptive weighting can scale to PointPillars with different fusion schemes and real-world datasets. Visualization examples further validate that the proposed weighting algorithm can mitigate the adverse effects under severe channel conditions and simultaneously enhance the received shared features when the channel condition improves.
\end{enumerate} 

The rest of the paper is organized as follows. Section \ref{related_work} will briefly review the related works. In Section \ref{sec2}, we will introduce the system model of the cooperative perception system and describe the backbone algorithm. Section \ref{sec3} will present the adaptive weighting algorithm and the self-supervised optimization. Simulation results will be demonstrated in Section \ref{sec4}. Finally, Section \ref{sec5} will conclude the work.

\section{Related work} \label{related_work}

\subsubsection{Single-vehicle perception}
Due to the advancement of sensor technologies and learning-based algorithms, the performance of 3D object detection for autonomous driving has been continuously improved using 3D scanners, such as light detection and ranging (LiDAR). Several approaches to processing the LiDAR point clouds have been proposed in \cite{8578570, s18103337, pointpillars, Pillar-Based, 8954080, 9157234}. The work in \cite{8578570} proposed a novel method to convert point clouds into 3D voxels and extract essential features. This detection algorithm adopts an end-to-end learning structure, enabling integration with a learning-based communications system for the end-to-end global optimization. To reduce the computational complexity, SECOND was proposed in \cite{s18103337}, where sparsely embedded convolutional layers were applied to 3D voxel features. On the other hand, the work in \cite{pointpillars} proposed to convert the voxels along the z-axis into pillars to avoid the computation of 3D voxels. Meanwhile, the imbalance issue caused by anchors was addressed in \cite{Pillar-Based}. Furthermore, a two-step detection framework was proposed in \cite{8954080}, where the rough estimation of proposals was generated first and refined in the second stage. In \cite{9157234}, the authors proposed to jointly leverage voxel-based and point-based detection methods. 

\subsubsection{Cooperative fusion}
Using the previous works on LiDAR-based 3D object detection backbones, efficient cooperative fusion schemes can be adopted for multiple CAVs. In \cite{8885377}, early fusion was first proposed to aggregate the raw point clouds shared among multiple CAVs, while the late fusion scheme was proposed in \cite{9228884} to use independently detected results for late fusion. 

To avoid the excessive bandwidth usage in early fusion, F-Cooper was proposed in \cite{10.1145/3318216.3363300} to use the intermediate features in cooperative perception. Furthermore, V2VNet in \cite{Wang2020V2VNetVC} used a graph neural network for the intermediate feature fusion considering the V2V communication with imperfect localization and time delay. To further reduce the bandwidth requirement, Who2com in \cite{9197364} proposed a three-stage handshake communication mechanism to select collaborators based on their matching scores. It was optimized by end-to-end supervision using the ground truth of the target task without annotations for the best agents to communicate with. Based on this handshake communication framework, When2com was proposed in \cite{9156848} to construct communication groups and learn when to communicate by using an asymmetric attention mechanism, while Where2comm \cite{where2comm} proposed a spatial-confidence-aware communication scheme to only transmit the crucial area of the perceptual information. In \cite{li2021learning}, a collaboration weighting graph was proposed for multi-vehicle perception, which requires fewer communication rounds than When2com and V2VNet. Furthermore, in \cite{9812038}, the attention mechanism was used to aggregate the intermediate features from CAVs, while the work in \cite{xu2022v2xvit} proposed a fusion framework based on vision Transformer to leverage the features from vehicles and infrastructures. Then, a hybrid object detection and tracking frame was proposed in \cite{10148929}, where historical tracking information was leveraged to enhance the inference for object detection with a spatial-temporal deep neural network. In \cite{XIA2023104120}, a data acquisition and analysis framework was proposed to collect and process the advanced sensor data from multiple CAVs for vehicle trajectory extraction, reconstruction, and evaluation, in which the Kalman filter and the Chi-square test were used to reduce the noise and outlier in the trajectories. In \cite{Qiao2023WACV}, a spatial-wise adaptive fusion was proposed for intermediate features. In \cite{10077757}, attention-based modules were adopted to enhance the interaction between the ego vehicle and other CAVs, and a repair network was used to mitigate the adverse effects of lossy communications. Most of these works focus on improving perception accuracy and reducing bandwidth utilization, assuming ideal communications. Very few have considered more realistic V2V communications, such as time delay \cite{xu2022v2xvit}, pose error \cite{Wang2020V2VNetVC, 9197364}, lossy communication \cite{10077757} and Rician fading with channel impairments \cite{10184097}. Among the existing works for cooperative fusion, \cite{li2021learning}, \cite{Qiao2023WACV}, and \cite{9197364} are more relevant to our work, where different weighting algorithms were investigated for fusion. However, these works are trained with well-annotated datasets for detection and neglect the realistic channel impairments.

\section{System model}\label{sec2}
\begin{figure}[!h] 
  \centering
  \includegraphics[width=3.0in]{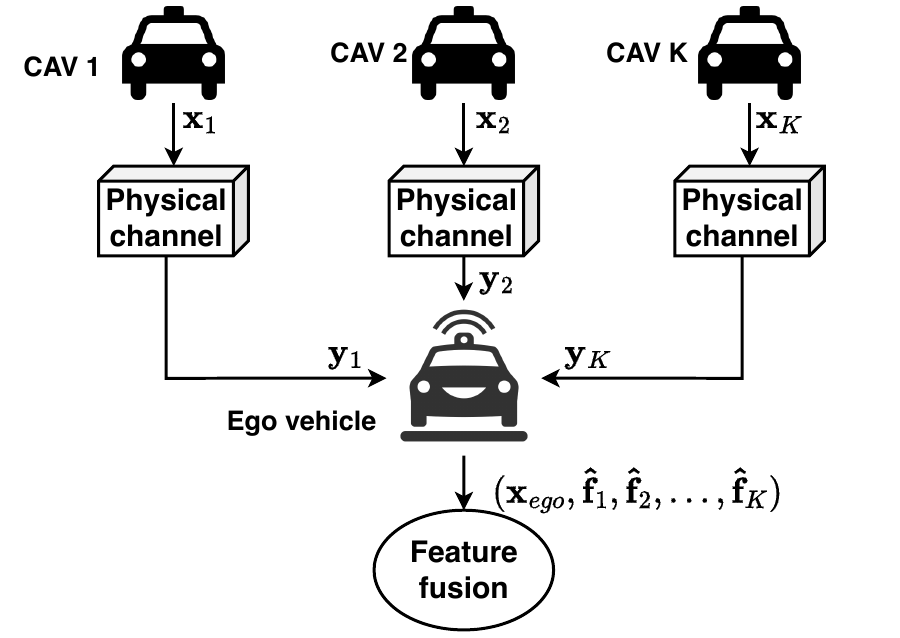}
  \caption{Cooperative perception via V2V communication.}
  \label{v2v_demo}
\end{figure}
\subsection{V2V communications channel model}
Consider a cooperative perception system where one ego vehicle is collaborating with $K$ CAVs. The shared information is transmitted from the CAVs to the ego vehicle, which is demonstrated in Fig. \ref{v2v_demo}.
\subsubsection{Rician fading}
Consider line-of-sight (LOS) signal propagation with path loss, the received signal at the ego vehicle from the $k$-th CAV is
\begin{equation}
    \mathbf{y}_k = \sqrt{\frac{p_0}{d_k^n}} h_k\mathbf{x}_k + \mathbf{w}_k, \label{eq1}
\end{equation}
where $\sqrt{\frac{p_0}{d_k^n}}$ is the path loss with parameter $p_0$ determined by antennas and channel characteristics, distance $d_k$ between transmitter and receiver, and path loss factor $n$, $h_k$ denotes the Rician fading channel following $\mathcal{CN}(\mu,\sigma_h^2)$, $\mathbf{y}_k \in \mathbb{C}^{L\times 1}$ denotes the complex-valued received signals from the $k$-th CAV, $\mathbf{x}_k \in \mathbb{C}^{L\times 1}$ is the transmitted signal, and $\mathbf{w}_k$ denotes the additive white Gaussian noise following $\mathcal{CN}(0,\sigma^2)$. 
\subsubsection{Multi-path}
In realistic communication, the communication channel may also suffer from multi-path fading. In addition to the communication model in (\ref{eq1}), a multi-path channel model using OFDM is also considered. The received symbols at $i$-th sub-carrier from the $k$-th CAV are 
\begin{equation}
    \boldsymbol{Y}_k[i] = \boldsymbol{H}_k[i]\boldsymbol{X}_k[i] + \boldsymbol{W}_k[i], \label{eq2}
\end{equation}
where $\boldsymbol{H}_k[i]$ denotes the channel frequency response, $\boldsymbol{W}_k[i]$ denotes the additive white Gaussian noise, and $\boldsymbol{X}_k[i]$ denotes the transmitted symbol. 


To cooperatively leverage the shared information from multiple CAVs, denote the aggregation of the shared information at the ego vehicle as 
\begin{equation}
    \mathbf{f}_{agg} = \mathcal{F}_{fusion}(\mathbf{f}_{ego},\mathbf{\hat{f}}_1,\mathbf{\hat{f}}_2,...,\mathbf{\hat{f}}_K ),
\end{equation}
where $\mathcal{F}_{fusion}(\cdot)$ denotes the fusion algorithm to combine the shared information received from CAVs, $\mathbf{f}_{ego}$ denotes the information sensed at the ego car itself, and $\mathbf{\hat{f}}_k$ denotes the shared information of the $k$-th CAV, reshaped from the recovered received signals $\mathbf{y}_k$ and $\boldsymbol{Y}_k$.

The sensed information from CAVs could provide informative viewpoints to improve the perception precision and range. However, when there are severe channel impairments, the recovered information $\mathbf{\hat{f}}_k$ suffers from distortion, thus compromising the aggregated feature $\mathbf{f}_{agg}$. In order to explore the effects of channel impairments, we consider the Rician fading channel with free-space path loss as in (\ref{eq1}) and the WINNER II channel model \cite{winner2} with multi-path fading as in (\ref{eq2}) to simulate the practical channel. Imperfect CSI with a Gaussian disturbance and pilot-based least-square channel estimation are adopted, accounting for channel estimation error. A zero-forcing detector is used to recover the signal $\mathbf{\hat{f}}_k$ from the received $\mathbf{y}_k$ or $\boldsymbol{Y}_k$. 

\subsection{Cooperative perception backbone}
\begin{figure*}[!h]
    \centering
    \includegraphics[width=\textwidth]{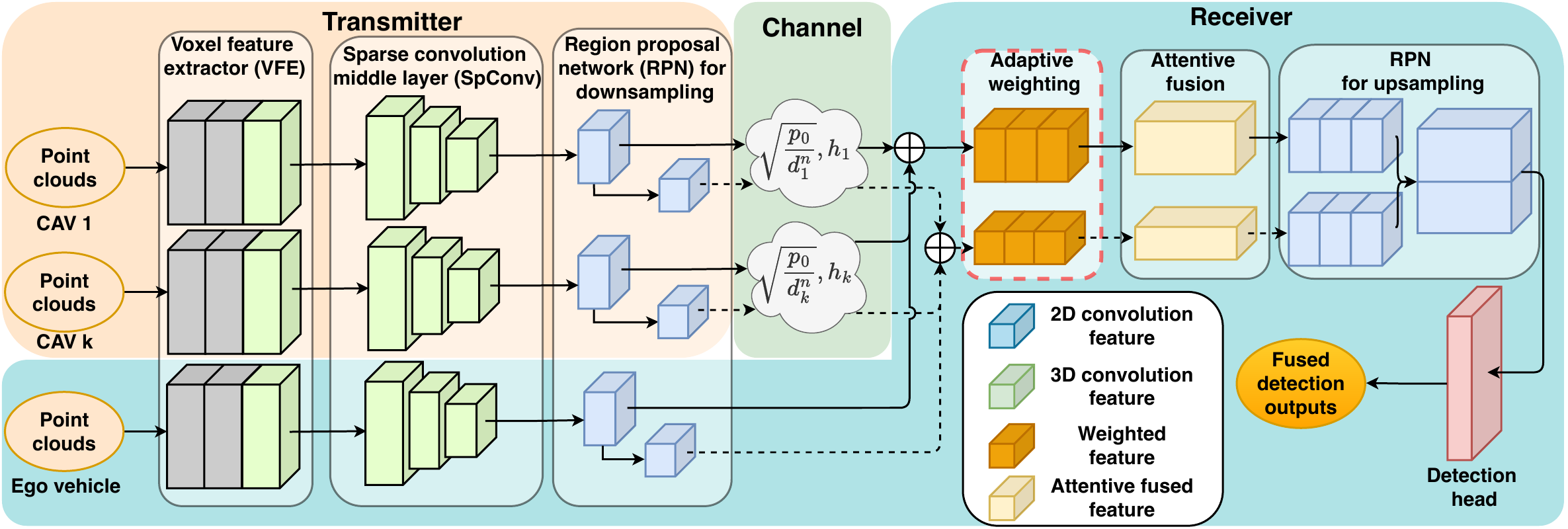}
    \caption{The system model of cooperative perception with learning-based communications system.}
    \label{weighting_system}
\end{figure*}
Fig. \ref{weighting_system} demonstrates the system model for cooperative perception with learning-based communications, the same as that in \cite{10184097}. SECOND \cite{s18103337} is adopted as the backbone algorithm for 3D detection. Based on the end-to-end learning structure of VoxelNet \cite{8578570}, SECOND incorporates sparsely embedded convolutional layers to improve the computational efficiency. This enables the integration with learning-based communications system for the end-to-end global optimization. SECOND has three main components voxel feature extractor (VFE), sparse convolution middle layer (SpConv), and region proposal network (RPN). Firstly, iterative conversion of raw point clouds into voxel representations is conducted by assigning the points to their corresponding voxels. Then, a VFE layer is employed to extract point-wise features from voxels. Subsequently, 3D sparse convolution is conducted for these point-wise features by only applying convolution to those non-zero elements of the sparse embeddings. This selective processing significantly reduces computational costs. The RPN adopts a residual structure, which involves downsampling and upsampling operations. To achieve downsampling, multiple layers of 2D convolution are applied to the features, along with batch normalization and rectified linear unit (ReLU) activation. On the other hand, upsampling is achieved by performing a 2D deconvolution on each level of downsampled features. Then, these deconvoluted features are concatenated to reconstruct the convolution feature, which serves as the input to the detection head. Finally, a single-shot detector (SSD) is used to output the classification results of objects and the regression results of their box localization.

To enable collaboration, multiple CAVs are connected to the ego vehicle. In this scenario, attentive fusion \cite{9812038} is employed as an intermediate fusion method to aggregate the received downsampled features. Attentive fusion process the features of each CAV via attention-based neural network. 

\subsection{Supervised end-to-end training} \label{supervised_training}
Supervised end-to-end training is adopted for cooperative perception with V2V communications, as the end-to-end trainable structure of SECOND allows any part of the information to be transmitted through the learning-based communications system. In this case, the learning-based communication and the detection algorithm can be trained together to consider signal distortion in communication using the loss function in \cite{s18103337, pointpillars}. Define the boxes and anchors for object detection as 
\begin{equation}
    (x, y, z, w, l, h, \theta), \label{coord}
\end{equation}
where $x$, $y$ and $z$ denote the coordinates of the box center, $w$, $l$ and $h$ denote the width, length and height of the box, $\theta$ denotes the rotation angle around z-axis. The localization residuals for elements in (\ref{coord}) can be expressed by
\begin{equation}
    \begin{split}
        &\Delta_x = \frac{x_{gt} - \hat{x}}{\hat{d}}, \Delta_y = \frac{y_{gt} - y_{pred}}{d}, \Delta_z = \frac{z_{gt} - z_{pred}}{\hat{d}},\\
        &\Delta_w = \log(\frac{w_{gt}}{w_{pred}}), \Delta_l=\log(\frac{l_{gt}}{l_{pred}}),  \Delta_h=\log(\frac{h_{gt}}{h_{pred}}),\\
        &\Delta_\theta = \sin(\theta_{gt} - \theta_{pred}),
    \end{split}
\end{equation}
where $\Delta$ is the residual between the ground truth and the prediction, the superscript $gt$ and $pred$ denote the ground truth and the prediction from the model, respectively. The regression loss of the box localization is computed by 
\begin{gather}
    \mathcal{L}_{reg}= \sum_{\lambda \in  (x,y,z,w,l,h,\theta)} \mathcal{F}_{SmoothL1}(\Delta_\lambda), \\
    \mathcal{F}_{SmoothL1}(x) = \begin{cases}
            0.5x^2 & \text{if} |x| < 1 \\
            |x|-0.5 & \text{otherwise}
        \end{cases}.
\end{gather}
To address the imbalance of the object samples in dataset, the focal loss in \cite{8237586} is adopted as the classification loss of target objects, which can be expressed by
\begin{equation}
    \mathcal{L}_{cls} = \mathcal{L}_{focal} = -\alpha(1 - q_{pred})^\gamma \log{(q_{pred})}
\end{equation}
where $q_{pred}$ is the model's estimated probability, $\alpha$ and $\gamma$ denote the parameters for the focal loss. Therefore, the overall loss can be calculated by
\begin{equation}
    \mathcal{L}_{total} = \frac{1}{N} (\beta_{reg}\mathcal{L}_{reg} + \beta_{cls}\mathcal{L}_{cls})  \label{supervised_loss}
\end{equation}
where $N$ denotes the number of positive anchors, $\beta_{reg}$ and $\beta_{cls}$ are the parameters for the regression loss and the classification loss, respectively. 

The cooperative perception model is trained by the Adam optimizer with adaptive learning rate and decay weight to minimize the total loss $\mathcal{L}_{total}$. This optimization process has been widely utilized by previous works as a solid framework for training an effective cooperative perception model, which will be adopted as the baseline for comparison in this work. 


\section{Adaptive weighting} \label{sec3}
In this section, the proposed CAV-level adaptive weighting algorithm will be discussed first. Then, the self-supervised training algorithm without labeled data will be presented.
\subsection{CAV-level adaptive weighting}
As shown in Fig. \ref{weighting_system}, the adaptive weighting is employed at the ego vehicle after receiving the features from other CAVs. This weighting algorithm is designed to exclude severely distorted information from certain CAVs or outliers in the fusion while leveraging the multiple viewpoints from other CAVs, with less distorted information having higher weights. This requires the weighting algorithm to generate weights for features from different CAVs. The weighted feature from the $k$-th CAV can be computed by
\begin{gather}
    \mathcal{W}_k = \mathcal{F}_{weighting}(\mathbf{f}_{ego},\mathbf{\hat{f}}_k) \\
    \mathbf{\tilde{f}}_{k} =  \mathcal{W}_k \mathbf{\hat{f}}_k
\end{gather}
where $\mathbf{f}_{ego}$ and $\mathbf{\hat{f}}_k \in \mathbb{R}^{B\times C\times H \times W}$ denote the intermediate features from the ego vehicle and the $k$-th CAV, respectively, $\mathcal{F}_{weighting}$ denotes the weighting algorithm, $\mathcal{W}_k \in \mathbb{R}^{B\times 1\times 1 \times 1} $ denotes the output weight and $\mathbf{\tilde{f}}_{k}$ denotes the weighted feature. 
\begin{figure}[!h] 
  \centering
  \includegraphics[width=2.3in]{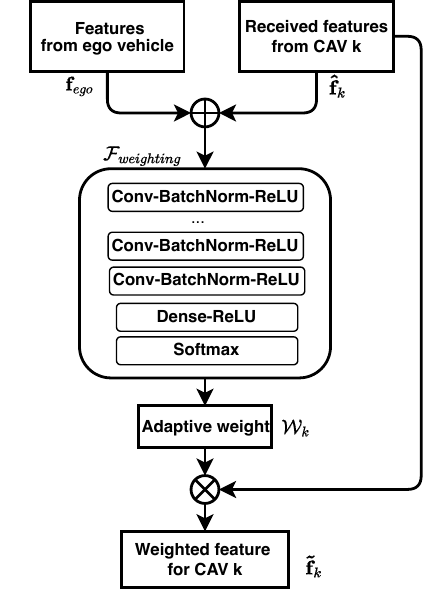}
  \caption{The illustration of adaptive weighting for the received features from other CAVs.}
  \label{weighting_pipeline}
\end{figure}
Fig. \ref{weighting_pipeline} demonstrates the working flow of the adaptive weighting for the received features from other CAVs. The input for the weighting algorithm is constructed by concatenating the features from the CAV and the ego vehicle. The underlying motivation behind this design is to effectively utilize the contrasting information in these two features. Despite having distinct viewpoints, the ego vehicle and CAV typically demonstrate similar data distribution and scales for the intermediate feature, particularly under optimal communication channel conditions. However, in scenarios where the communication channel is severely corrupted, there is a substantial difference between the CAV feature and the ego vehicle feature. Therefore, the contrasting characteristics between these two features could offer valuable insights for the weighting algorithm to generate adaptive weights. 

In order to exploit these contrasting features, we use 2D CNN combined with batch normalization (BatchNorm) and ReLU activation to process the input. Multiple layers of CNN-BatchNorm-ReLU could enhance feature learning and allow for effective gradient propagation through deep networks. Moreover, BatchNorm and ReLU are essential due to its capability of preventing gradient vanishing. Instead of generating pixel-wise and channel-wise fusion weights, as in \cite{li2021learning} and \cite{Qiao2023WACV}, we adopt CAV-level weighting that generates a value between 0 to 1 for each CAV for computational efficiency, which is similar to Who2com \cite{9197364} but focuses on mitigating the channel distortion. To achieve this, we flatten the multi-dimensional latent features processed by the CNN-BatchNorm-ReLU blocks. Then, linear dense layers and ReLU are used to facilitate fast convergence and mitigate the gradient vanishing. For the output layer, Softmax is employed to generate the regression results instead of using Sigmoid, as Sigmoid leads to severe gradient vanishing in self-supervised models without ground truth labels. Softmax is often utilized in multi-classification tasks to produce a probability distribution where the sum of the probabilities of all classes equals 1. In this work, we define two classes representing the positive or negative impact on the performance. The probability of the positive class is used as the output weight for the feature of the CAV. 

\subsection{Self-supervised optimization} \label{3.2}
The primary purpose of this weighting model is to mitigate the negative effect of channel distortion when it is severe, while minimizes its influences when the channel conditions are good. This requires the weighting model to adapt to the shared information with different levels of communication channel distortion. Unlike the works in \cite{li2021learning}, \cite{Qiao2023WACV}, and \cite{9197364}, which rely on the supervision of annotated datasets to determine weights for cooperative perception, we adopt self-supervised training that only requires the shared information and its augmentations for scalability and data efficiency. Specifically, a tailored self-supervised loss to optimize the weighting model $\mathcal{F}_{weighting}$ is proposed, which can be expressed by

\begin{gather}
    \mathcal{L}_{self} = \frac{1}{K} (\lambda_{pos}\mathcal{L}_{pos} + \lambda_{neg}\mathcal{L}_{neg} ), \label{self_loss}\\
   \mathcal{L}_{pos} =\sum_{k=1}^K \mathcal{D}_{kl}[\mathcal{S}(\mathcal{W}^{+}_k\mathbf{f}^{+}_{k}) || \mathcal{S}(\mathbf{f}_k)], \label{pos_loss}\\
    \mathcal{L}_{neg} =\sum_{k=1}^K \mathcal{D}_{kl}[ \mathcal{S}(\mathcal{W}^{-}_{k} \mathbf{f}^{-}_{k}) ||  \mathcal{S}(\mathbf{f}_k)],  \label{neg_loss}
\end{gather}
where $K$ is the number of CAVs, $\mathcal{L}_{pos}$ and $\mathcal{L}_{neg}$ denote the positive loss and negative loss with hype-parameter $\lambda_{pos}$ and $\lambda_{neg}$, respectively, $\mathcal{S}$ denotes the Softmax function, $\mathcal{D}_{kl}(\cdot)$ denotes the Kullback–Leibler (KL) divergence, $\mathbf{f}^{+}_{k}$ and $\mathbf{f}^{-}_{k}$ denote the positive and negative augmentation experiencing minor and critical channel distortions, respectively, and $\mathcal{W}^{+}_k$ and $\mathcal{W}^{-}_{k}$ denote the corresponding adaptive weights.

KL divergence enables the comparison between two probability distributions with different scales, offering a means to quantify the extent to which the distribution of the weighted feature deviates from that of the transmitted feature. \hlbreakable{Instead of depending on external ground-truth annotations for object detection as in supervised learning, self-supervised learning leverages the shared information itself to generate internal supervisory signals for training models. By simulating light and severe channel distortions for the positive and the negative augmentation, respectively, their contrastive information can enhance the adaptability to various channel conditions. Specifically, minimizing both positive and negative loss in the same iteration allows the weighting model to generate adaptive weights for features with different distortion levels. Ideally, the output weight for the less distorted feature should be close to 1, while the output for the severely polluted features should be close to 0 to minimize its negative effects. Furthermore, the self-supervised method is independent of the supervised training, allowing for a flexible extension to diverse communication environments without any modifications to the backbone network.}

\hlbreakable{\subsection{Training strategy}}
\hlbreakable{Three training schemes are considered to evaluate the performance of cooperative perception with V2V communications. \textit{Training Scheme 1} assumes the ideal communication for the cooperative perception without an adaptive weighting module, which is widely used in the benchmarks in \cite{9812038,10203124}. Different from \textit{Training Scheme 1} that ignores channel distortion in training, \textit{Training Scheme 2} adopts a distortion-in-the-loop training strategy by incorporating a communication channel model into the cooperative perception system. Since we cannot assume the realistic channels in the training stage, this communication channel is usually mathematically simulated. In this case, the cooperative perception is trained in an end-to-end manner by using the supervised loss function in (\ref{supervised_loss}) and applying a simulated distortion to the shared information to enhance the model robustness, which is adopted in \cite{10184097}. }


\hlbreakable{Unlike \textit{Training Scheme 1} and \textit{2} focus on the supervised optimization for the cooperative perception without weighting module, the proposed weighting module is introduced and optimized by \textit{Training Scheme 3}. One prerequisite for this optimization is a pre-trained cooperative perception by \textit{Training Scheme 2}, which generates the shared information for self-supervised training. However, the ground-truth annotations for the detection tasks are not required for training the adaptive weighting module. The proposed adaptive weighting is trained solely with the self-supervised loss in (\ref{self_loss}) using the shared information and its simulated augmentations, while the parameters of this pre-trained model will not be updated at this stage. The purpose of this design is to dynamically address the variations of channel distortion by adaptive weighting without affecting the cooperative detection systems. Additionally, the self-supervised training without annotated datasets contributes to data and computation efficiency, enabling flexible adaption to diverse communications environments.}

\hlbreakable{In this work, the performance of different training schemes will be evaluated in Section \ref{sec4.4}, and the training details will be given in Section \ref{sec4.1}. \textit{Training Scheme 2} will be adopted as benchmarks without adaptive weighting for comparison in other scenarios in Section \ref{sec4}. }

\section{Numerical results and discussion} \label{sec4}
In this section, the simulation settings will be presented first. Then, the proposed adaptive weighting model will be evaluated for various effects. 

\subsection{Simulation settings} \label{sec4.1}
The simulation settings are as below:
\subsubsection{Dataset}To conduct the training and evaluation, we utilize the OPV2V dataset proposed in \cite{9812038}, which is constructed using OpenCDA simulation tool \cite{10045043}. OPV2V consists of the default CARLA towns and the Culver city dataset. The default CARLA towns consists of 6,765 samples for training and 1980 samples for validation, while the Culver city has 550 samples to evaluate the domain adaptability of the proposed model as the test set for generalization ability. In addition to the simulation dataset, V2V4Real dataset \cite{10203124} is adopted as the real-world dataset, which is collected by two vehicles considering diverse real-world scenarios.

\subsubsection{Baseline}Attentive fusion \cite{9812038} with SECOND detection backbone \cite{s18103337} is employed as the baseline method for the cooperative perception. As a benchmark, the single-vehicle perception using only the sensed information from the ego car itself is adopted. Additionally, PointPillars \cite{pointpillars} with attentive fusion \cite{9812038} and V2VNet \cite{Wang2020V2VNetVC} are implemented to evaluate the scalability of the proposed methods. Average precision (AP) is adopted as the performance metric, which computes the average precision according to the recall value at different thresholds of intersection over union (IoU).

\subsubsection{Communication settings}Rician fading with Rician $K$ factor of 1 and free space path loss are considered to simulate the LOS channel. A Gaussian disturbance with mean 0 and variance 0.1 is added to the CSI to simulate the imperfect CSI for the Rician fading channel. Additive white Gaussian noise is considered with the signal-to-noise ratio (SNR) from -10 dB to 30 dB. Moreover, various path loss factors from 1 to 3 are also considered. \hlbreakable{To account for the realistic communication channel, we use the WINNER II channel\cite{winner2} in an OFDM system with 64 sub-carriers to consider the multi-path effects.} The carrier frequency is 2.6 GHz, the number of paths is 24, and the maximum delay is 16. To consider the channel estimation error, the least-square channel estimator with different pilots is also taken into account. \hlbreakable{It is important to note that the simulated Rician fading channel is used in training, while the WINNER II channel is only applied in inference to validate the performance of the proposed method as unseen realistic distortions. }

\subsubsection{Training} \hlbreakable{For \textit{Training Scheme 1}, ideal communication is considered for the cooperative perception without adaptive weighting module, the same as the benchmarks in \cite{9812038, 10203124}. For \textit{Training Scheme 2}, a Rician fading channel with an SNR at 15 dB is simulated for the distortion-in-the-loop training, adopted as the baseline for cooperative perception without weighting. The reason for adopting a mild distortion in this training is to enhance the model's robustness to distortions while preventing the overfitting on the patterns of the distorted samples. To train the proposed adaptive weighting by \textit{Training Scheme 3}, the pre-trained model by \textit{Training Scheme 2} is used to generate the shared information without updating its parameters. Then, a mathematically simulated Rician fading channel with an SNR of 30 dB and -10 dB is applied to the shared information to generate the positive augmentation $\mathbf{f}^{+}_{k}$ in (\ref{pos_loss}) and the negative augmentation $\mathbf{f}^{-}_{k}$ in (\ref{neg_loss}), respectively. Lastly, the proposed adaptive weighting is trained solely with the self-supervised loss function in (\ref{self_loss}) based on the shared information and augmentations, where $\lambda_{pos}$ is 1 and $\lambda_{neg}$ is 0.0001.}

For the proposed adaptive weighting model, we use 4 layers of CNN-BatchNorm-ReLU blocks to process the inputs, and then flatten the convoluted features by one layer of Dense-ReLU block. Finally, a Softmax is used to produce a probability distribution for two classes, positive and negative. The probability of the positive class is used as the output weight for the feature of the CAV. The computational resource for the implementation, training and evaluation of the cooperative perception is provided by Oracle Cloud Infrastructure with NVIDIA Tesla V100 GPUs.

\subsection{Performance in Rician fading channel}\label{sec4.2}
\begin{figure}[!ht] 
    \centering
    \subfloat[]{\includegraphics[width=3.2in]{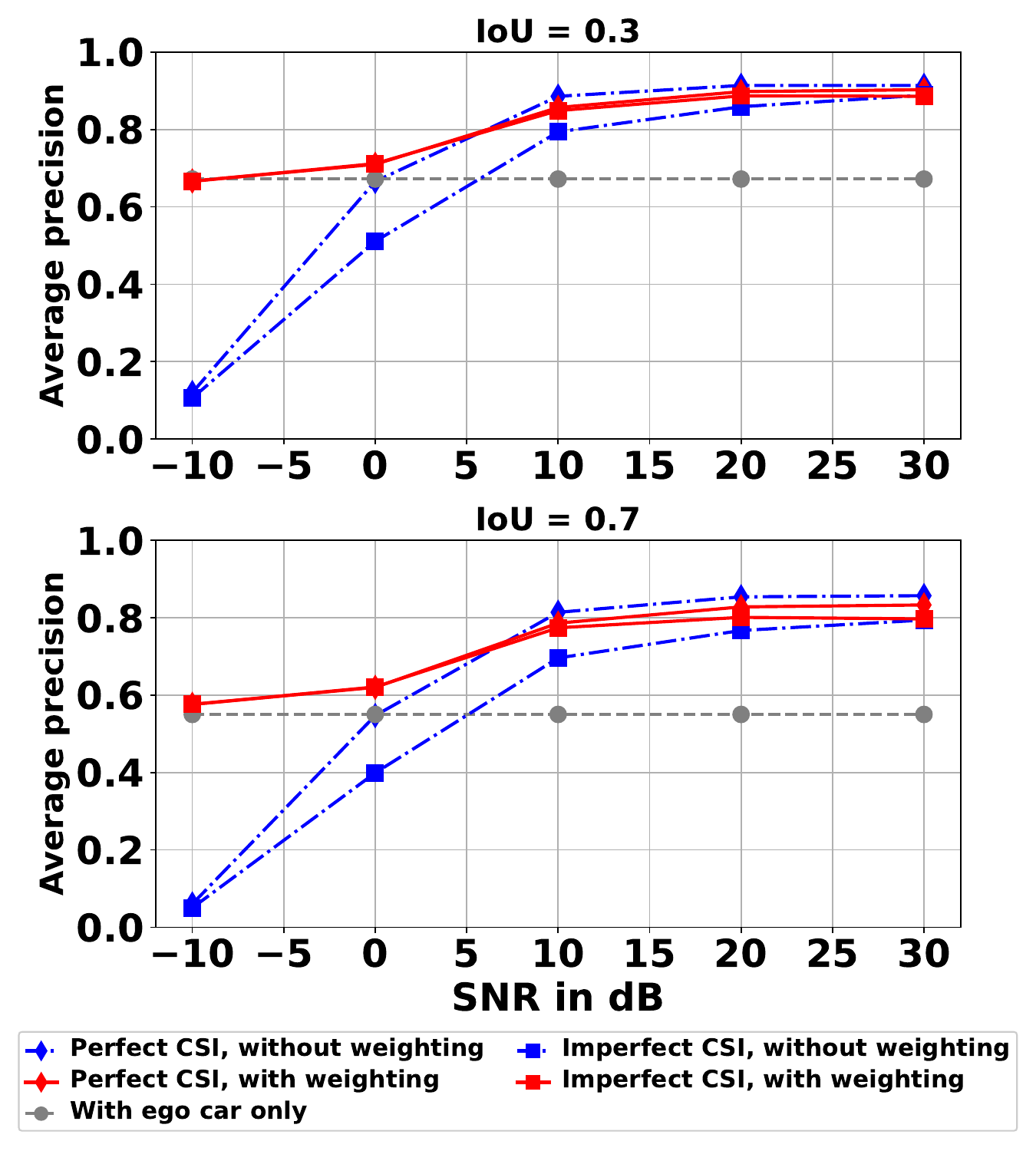}
    \label{fading.default}}
    \hfil
    \subfloat[]{\includegraphics[width=3.2in]{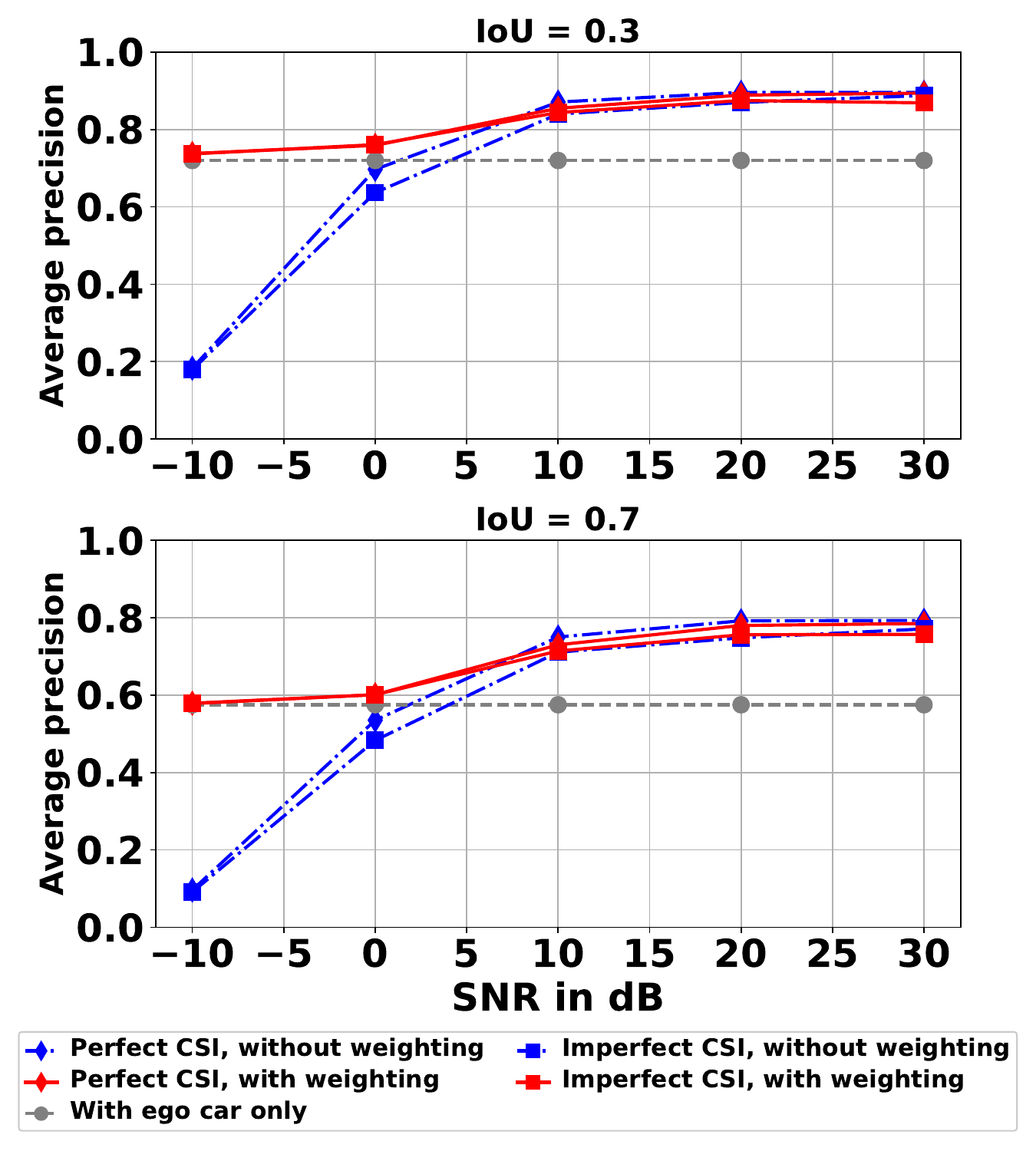}
    \label{fading.culver}}
    \caption{Average precision with Rician fading, free path loss and noise. (a) Default towns. (b) Culver city.}
    \label{fig:fading}
\end{figure}

\begin{figure}[!ht] 
    \centering
    \subfloat[]{\includegraphics[width=3.2in]{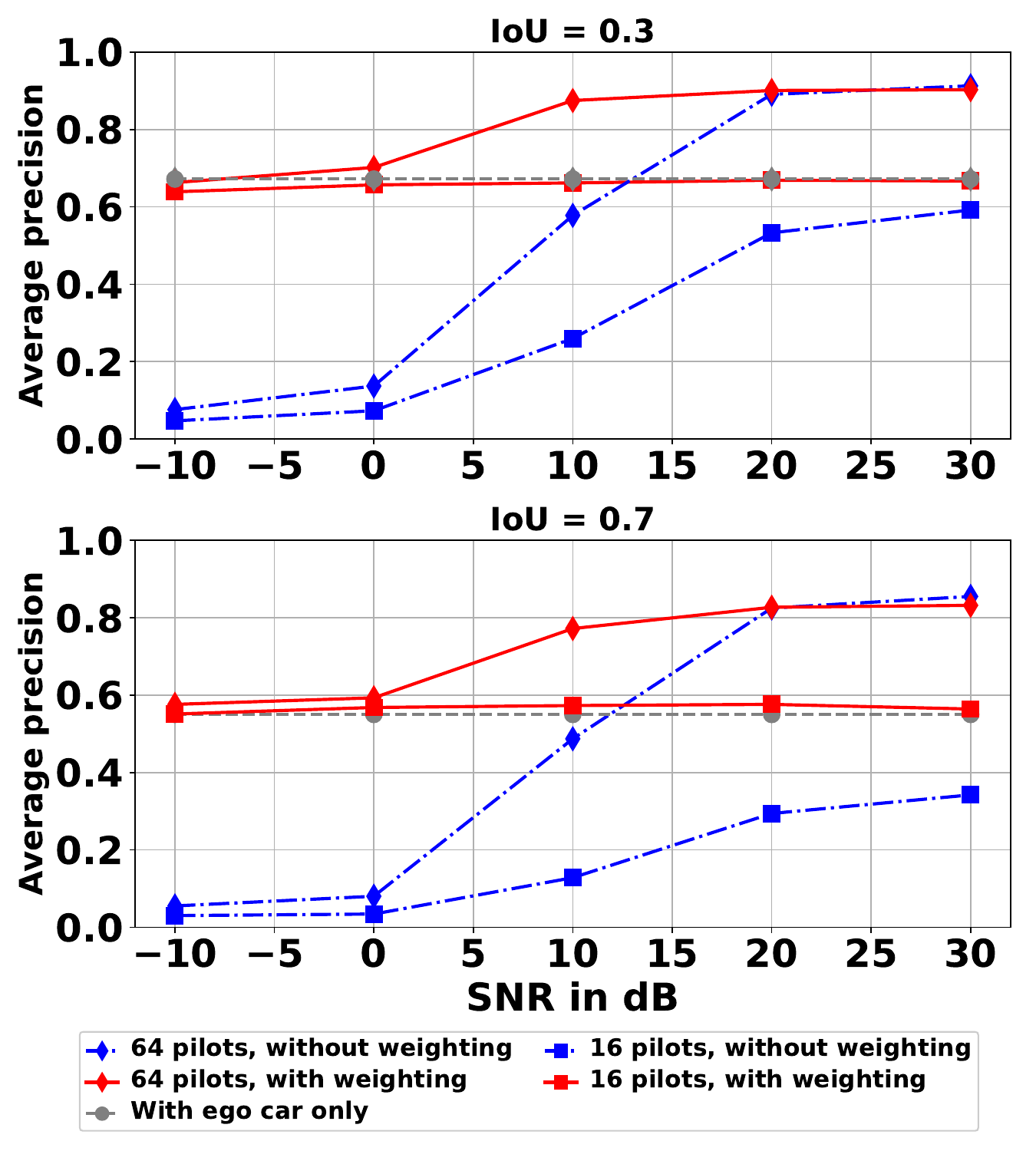}
    \label{ofdm.default}}
    \hfil
    \subfloat[]{\includegraphics[width=3.2in]{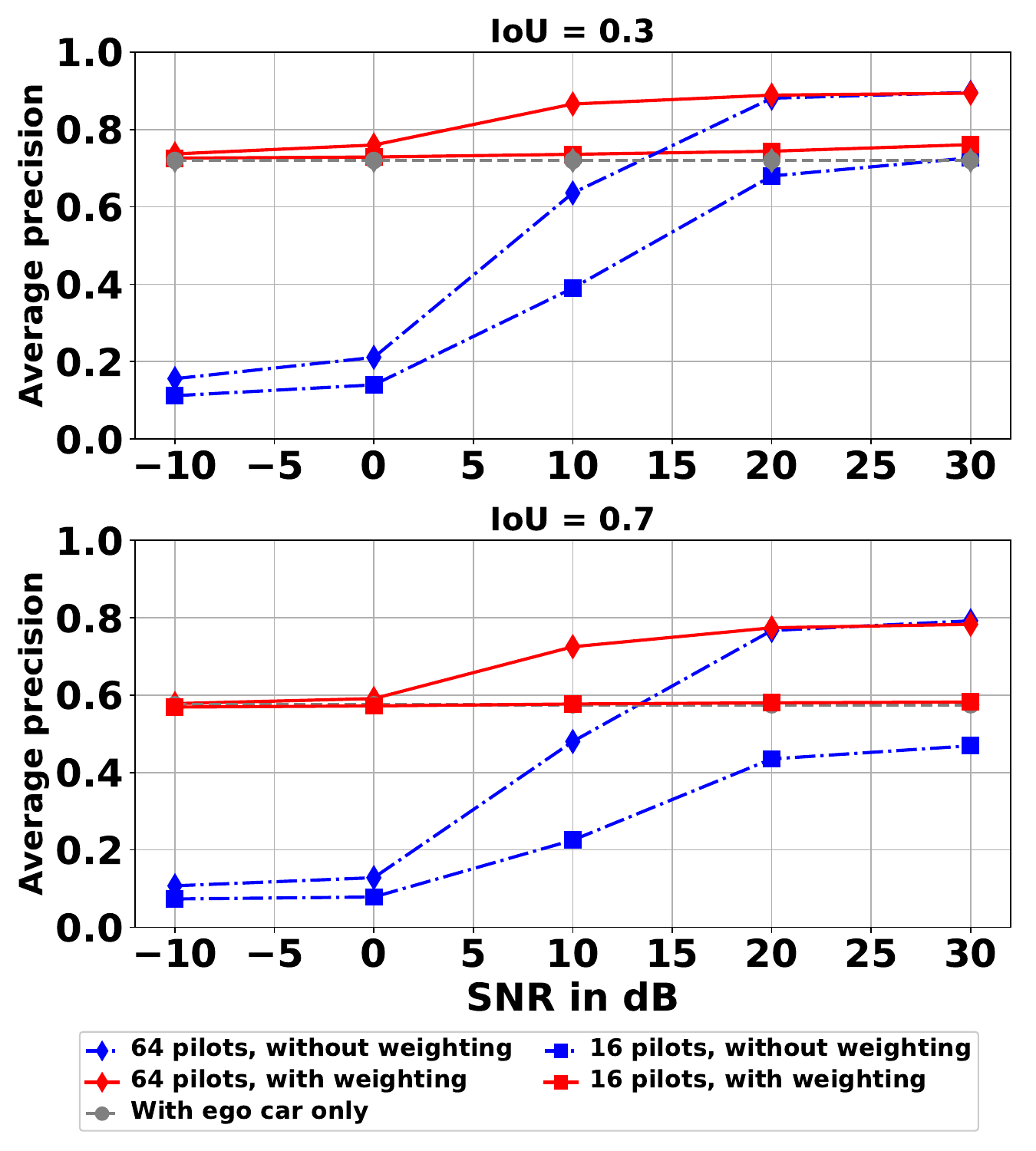}
    \label{ofdm.culver}}
    \caption{Average precision with WINNER II channel. (a) Default towns. (b) Culver city.}
    \label{fig:ofdm}
\end{figure}

Fig. \ref{fig:fading} shows the performance of cooperative perception in the presence of Rician fading, path loss and noise, where both perfect and imperfect CSI are considered. 

In Fig. \subref{fading.default}, we evaluate the performance on the dataset of the default towns. For the cooperative perception without adaptive weighting, it experiences a significant decline in average precision from approximately 85\% to 5\% for IoU=0.7 and from 90\% to 10\% for IoU=0.3, when the SNR decreases from 30 to -10 dB, assuming the perfect CSI. This is because the fusion is degraded by outliers with very noisy shared information. For the single-vehicle perception, it has stable performance at about 60\% for IoU=0.7 and 65\% for IoU=0.3 regardless of the SNRs, as it only uses its own sensed information at the ego car to avoid the distortion incurred in shared information. However, this method cannot leverage the multiple viewpoints when the channel condition improves. The cooperative perception with adaptive weighting could mitigate the adverse effects of severe channel impairments and leverage the shared information when there is limited channel distortion. Specifically, it achieves an accuracy at about 60\% for IoU=0.7 and 65\% for IoU=0.3 when the SNR decreases to -10 dB while having a similar performance to the non-weighted cooperative perception when the SNR is larger than 10 dB. These observations demonstrate that the weighting model outperforms the single-agent perception and the non-weighted cooperation by intelligently adapting to different levels of channel impairments. 

Furthermore, obtaining the perfect CSI is not practical in realistic scenarios; thus, a Gaussian disturbance is added to CSI to simulate the imperfect CSI. Without weighting, the cooperative perception experiences around 10\% performance degradation due to the imperfect CSI. However, with the implementation of adaptive weighting, the performance is nearly unaffected by imperfect CSI because of the effective mitigation of the negative effects. In this case, the system aided by adaptive weighting outperforms the non-weighted cooperation and single-agent perception across all SNRs. 

Fig. \subref{fading.culver} shows the performance using the dataset of the Culver city. Similar to the performance on the default dataset, the weighted cooperative perception performs better than the non-weighted systems when the SNR is less than 10 dB. When the SNR is larger than 10 dB, the perception with weighting achieves a very close accuracy to the baseline without weighting, which outperforms the single-agent perception. This also validates that the proposed weighting model is well-trained and adaptive to datasets from diverse domains.

\subsection{Performance in multi-path fading channel}\label{sec4.3}
Fig.\ref{fig:ofdm} shows the performance of cooperative perception in WINNER II channel. An OFDM system with 64 sub-carriers is adopted. Pilot-based least-square channel estimation is used to account for the channel estimation error.  

In Fig. \subref{ofdm.default}, when 16 pilots are used and the SNR is 30 dB, the baseline without weighting can only achieve an accuracy at approximately 30\% for IoU=0.7 and 60\% for IoU=0.3. However, stable performance at about 60\% for IoU=0.7 and 65\% for IoU=0.3 can be obtained by the proposed weighting model across all SNRs, which is almost identical to the performance with ego vehicle only but significantly outperforms the one without weighting.

If 64 pilots are used, the weighted cooperative perception performs better than the non-weighted model when the SNR increases from -10 to 20 dB while maintaining a similar performance to the non-weighted cooperation for the SNR over 20 dB. Compared with the single-agent detection without collaboration, the non-weighted system can obtain a better accuracy only when the SNR is over 20 dB. In contrast, the weighted system performs better regardless of the SNRs. This demonstrates that the weighting algorithm works effectively on the realistic channel with multi-path fading. Furthermore, since the WINNER II channel is not used to train the weighting model, the proposed weighting model is also proved to generalize well to unseen communications channels. Similar observations can be made for the Culver city dataset in Fig. \subref{ofdm.culver}.  
\hlbreakable{
\subsection{Performance for different training schemes}}\label{sec4.4}
\begin{table}[!ht]
\caption{Performance of cooperative perception for different training schemes with WINNER II channel. (a) Default towns. (b) Culver city.}
\label{table1}
\centering
\setlength\tabcolsep{ 1pt}
    \subfloat[]{
    \label{table1.a}
    \renewcommand{\arraystretch}{1.2}
    \centering
    \setlength\tabcolsep{ 1pt}
    \hlbreakable{
    \begin{tabular}{@{ }c@{ }|cccccc@{ }}
        \toprule
         \multirow{2}{*}{SNR} & \multicolumn{2}{c}{\begin{tabular}[c]{@{ }c@{ }} \textit{Training Scheme 1} \end{tabular}} & \multicolumn{2}{c}{\begin{tabular}[c]{@{ }c@{ }} \textit{Training Scheme 2} \end{tabular}} & \multicolumn{2}{c}{\begin{tabular}[c]{@{ }c@{ }} \textit{Training Scheme 3} \end{tabular}} \\
         &AP@0.3&AP@0.7&AP@0.3&AP@0.7&AP@0.3&AP@0.7\\ \midrule
        Ideal& 0.899& 0.831& \textbf{0.914}&\textbf{0.857} & 0.903& 0.833\\
         30 dB& 0.857& 0.786& \textbf{0.913} & \textbf{0.855} & 0.903 & 0.832\\
         10 dB& 0.089& 0.054& 0.578& 0.487& \textbf{0.875}& \textbf{0.772}\\
        -10 dB&0.042 &0.023 &0.076 &0.055 &\textbf{0.663} &\textbf{0.576} \\ \bottomrule
    \end{tabular}%
    }
}
\hfill    
    \subfloat[]{
    \renewcommand{\arraystretch}{1.2}
    \label{table1.b}
    \centering
    \hlbreakable{
        \begin{tabular}{@{ }c@{ }|cccccc@{ }}
            \toprule
             \multirow{2}{*}{SNR}& \multicolumn{2}{c}{\begin{tabular}[c]{@{ }c@{ }}\textit{Training Scheme 1}\end{tabular}} & \multicolumn{2}{c}{\begin{tabular}[c]{@{ }c@{ }}\textit{Training Scheme 2}\end{tabular}} & \multicolumn{2}{c}{\begin{tabular}[c]{@{ }c@{ }}\textit{Proposed Method} \end{tabular}} \\
             &AP@0.3&AP@0.7&AP@0.3&AP@0.7&AP@0.3&AP@0.7\\ \midrule
            Ideal& 0.894& 0.754& \textbf{0.897}&\textbf{0.794} & 0.894& 0.783\\
            30 dB& 0.862& 0.720& \textbf{0.894} & \textbf{0.791} & \textbf{0.894} & 0.783\\
            10 dB& 0.133& 0.075& 0.641& 0.488& \textbf{0.866}& \textbf{0.725}\\
            -10 dB&0.094 &0.057 &0.154 &0.107 &\textbf{0.737} &\textbf{0.578} \\ \bottomrule
        \end{tabular}%
        }
    }
\end{table}

\hlbreakable{In this section, we compare the performance of cooperative perception using different training schemes. \textit{Training Scheme 1} uses ideal communication for training, and \textit{Training Scheme 2} adopts a distortion-in-the-loop training strategy. \textit{Training Scheme 1} and \textit{2} do not have an adaptive weighting module, while \textit{Training Scheme 3} additionally applies adaptive weighting to a pre-trained model based on distortion-in-the-loop training. The performances are validated on the WINNER II channel with an SNR of 30 dB, 10 dB, and -10 dB for the unseen realistic channel with light, mild, and severe distortions, respectively.}

\hlbreakable{Table \subref{table1.a} presents the average precision of different training schemes on the default town dataset. \textit{Training Scheme 1} can barely achieve an accuracy of 10\% when the SNR is 10 dB and -10 dB. On the other hand, \textit{Training Scheme 2} outperforms \textit{Training Scheme 1} in all communication conditions, demonstrating enhanced robustness to channel distortions. Even in ideal communication scenarios, \textit{Training Scheme 2} performs better than \textit{Training Scheme 1}. This is because the simulated light distortion on the shared information serves as a form of data augmentation, improving the system robustness and avoiding overfitting on the training samples. However, when there is severe channel distortion, the performance of \textit{Training Scheme 2} experiences a significant degradation to below 10\%. To address this, \textit{Training Scheme 3} obtains the best accuracy among these three schemes when the SNR is 10 dB and -10 dB. This underscores the efficacy of \textit{Training Scheme 3} in mitigating the adverse effects of severe channel distortion. Meanwhile, although \textit{Training Scheme 3} has a very slight performance degradation compared with \textit{Training Scheme 2} when the channel condition improves, \textit{Training Scheme 3} can achieve over 90\% for IoU=0.3 and 80\% for IoU=0.7, outperforming \textit{Training Scheme 1}. This observation highlights the adaptability of \textit{Training Scheme 3} to the non-distorted communications scenarios, maintaining competitive performance to models trained under ideal communications and simulated distortions. Similar observations and conclusions can be obtained for the performance on the Culver city dataset in Table \subref{table1.b}, which validates the effectiveness of the proposed adaptive weighting model.}

\subsection{Performance for different path loss factors}
\begin{figure}[!ht] 
    \centering
    \includegraphics[width=3.0in]{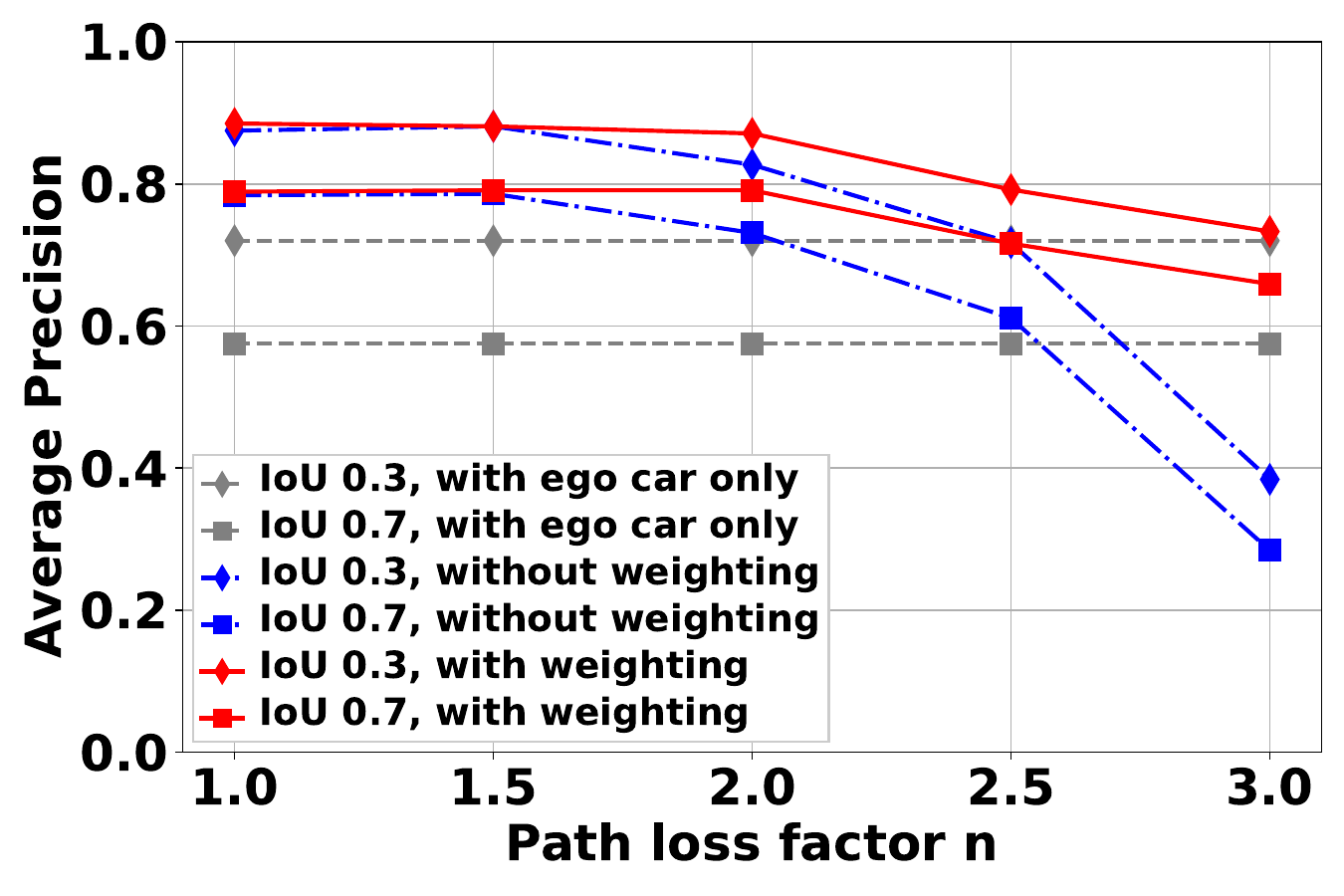}
    \caption{Average precision with different path loss factor $n$.}
    \label{fig:pathloss}
\end{figure}
Fig. \ref{fig:pathloss} depicts the performance curves of the cooperative perception with various path loss factors when the SNR is 30 dB with imperfect CSI. When the path loss factor increases from 1 to 3, the non-weighted cooperative perception experiences a substantial decline in accuracy, decreasing from about 80\% to 30\% for IoU=0.7 and from 90\% to 40\% for IoU=0.3, respectively. For the cooperative perception with adaptive weighting, the accuracy ranges from approximately 80\% to 65\% for IoU=0.7 and 88\% to 73\%, which is more robust than the non-weighted system. Moreover, when the path loss factor ranges from 2 to 3, the weighted model performs better than both baselines with non-weighted cooperation and single-agent perception. Therefore, the proposed weighting model can adapt to different communications environments with different path loss factors and mitigate its negative effects. This is important for autonomous driving, as the car can move from city centre to highway with different $n$.

\subsection{Performance on real-world datasets using other backbones} \label{sec4.5}
\begin{figure}[!ht] 
    \centering
    \includegraphics[width=3.2in]{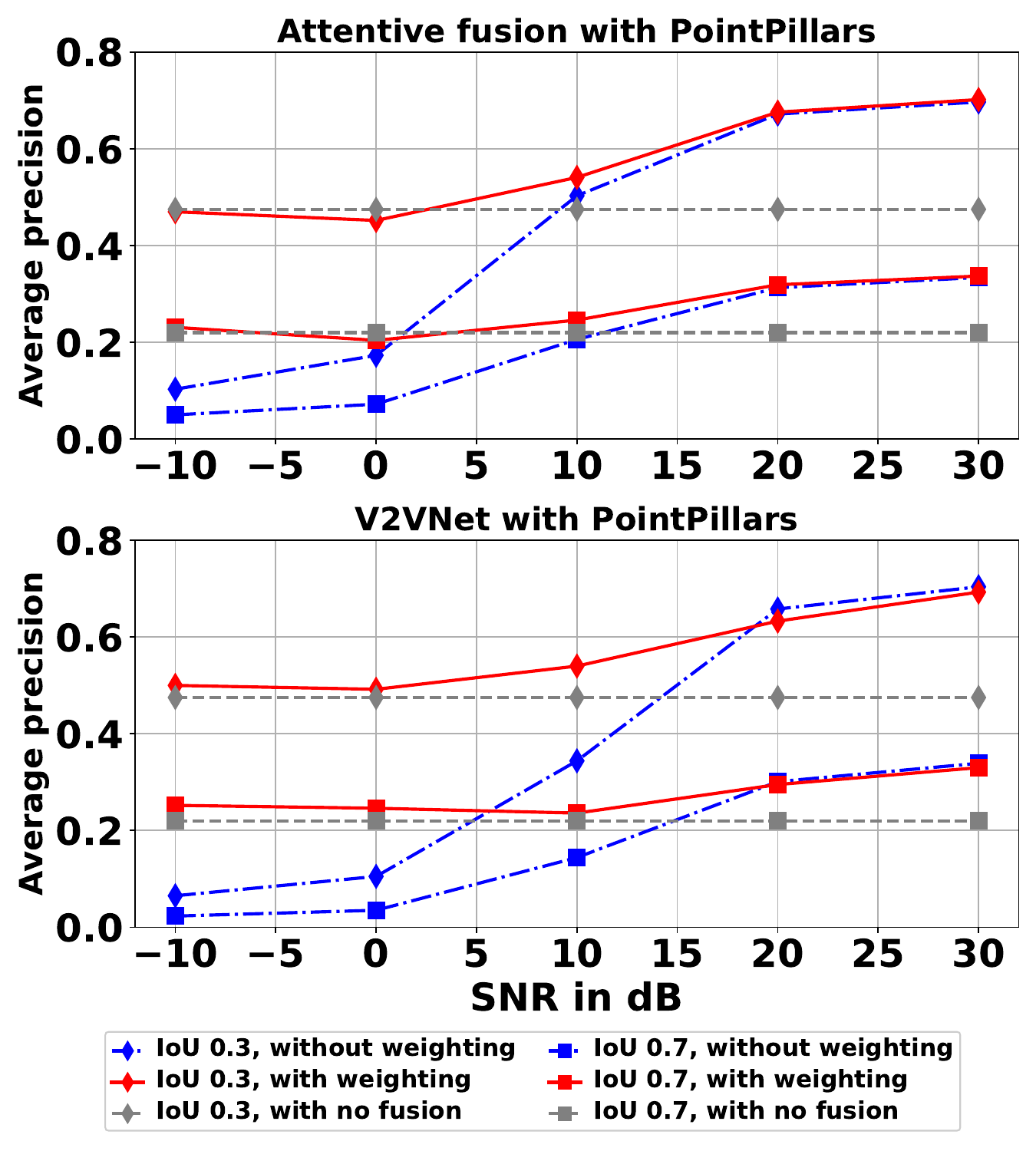}
    \caption{Performance of the attentive fusion and V2VNet using the PointPillars backbone on the V2V4Real dataset in the WINNER II channel. }
    \label{fig:v2v4real}
\end{figure}

\begin{figure*}[!h] 
    \centering
    \subfloat[]{\includegraphics[width=3.3in]{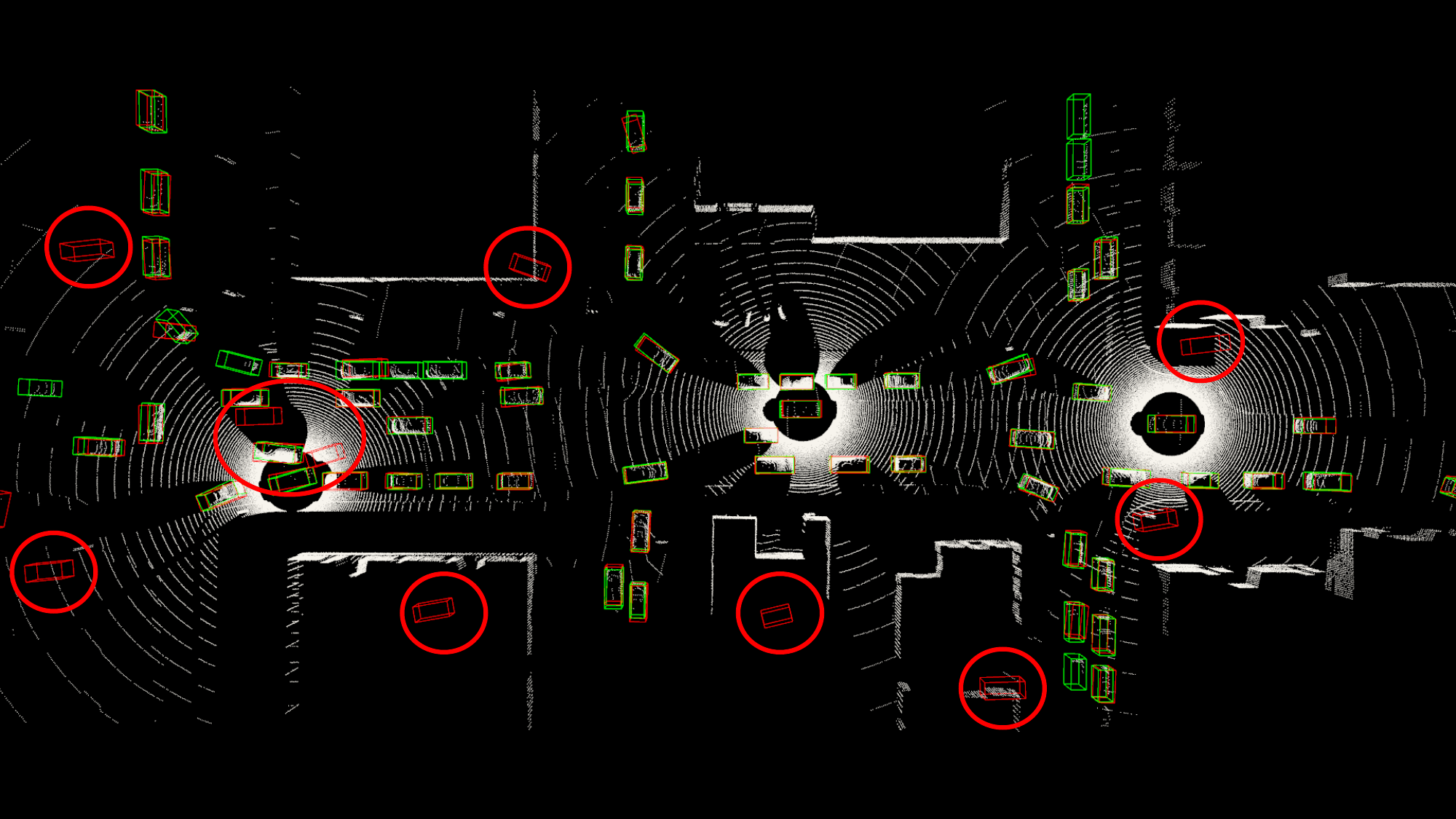}
    \label{ofdm.vis1}}
    \hfil
    \subfloat[]{\includegraphics[width=3.3in]{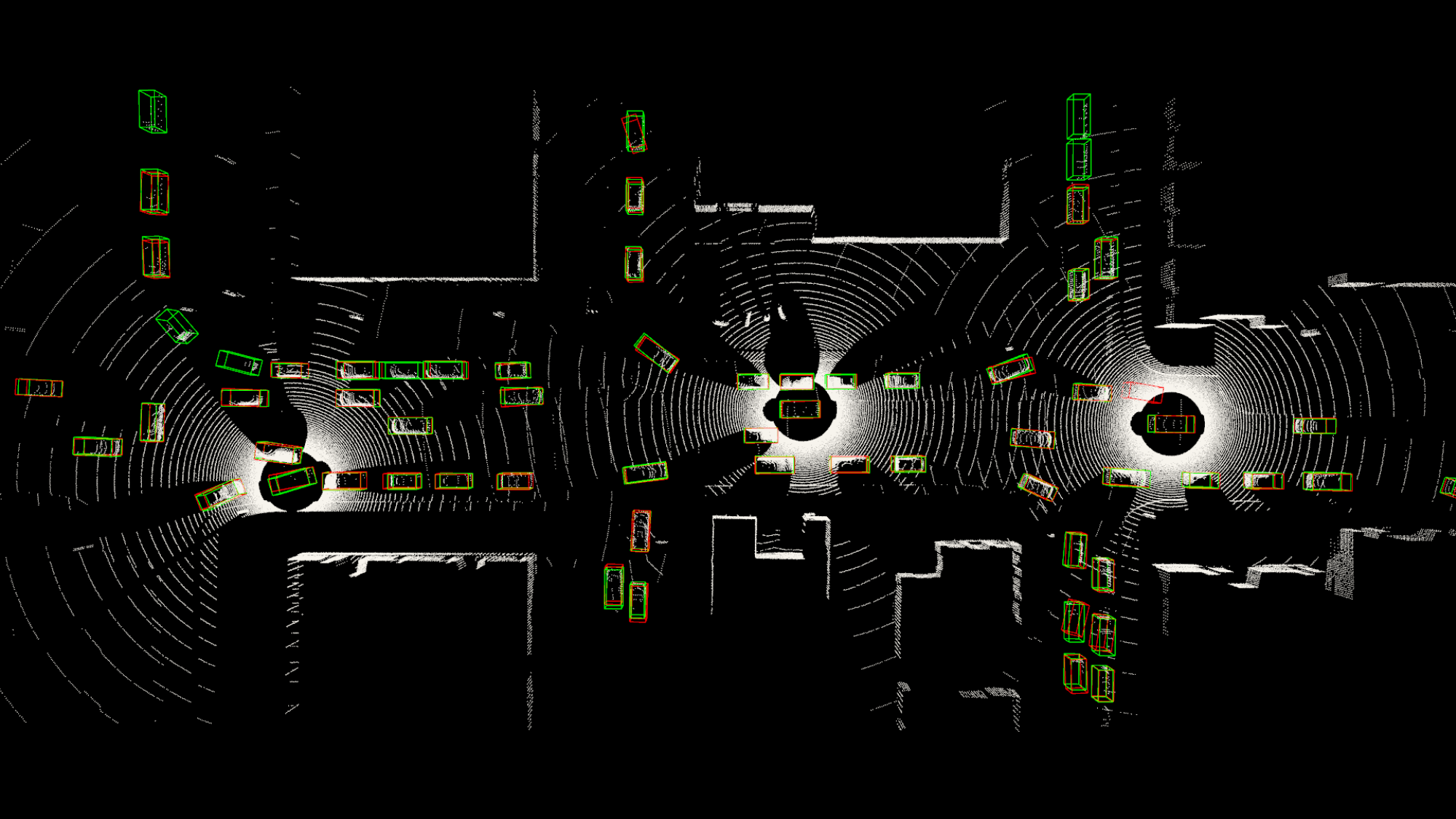}
    \label{ofdm.vis2}}
    \hfil
    \subfloat[]{\includegraphics[width=3.3in]{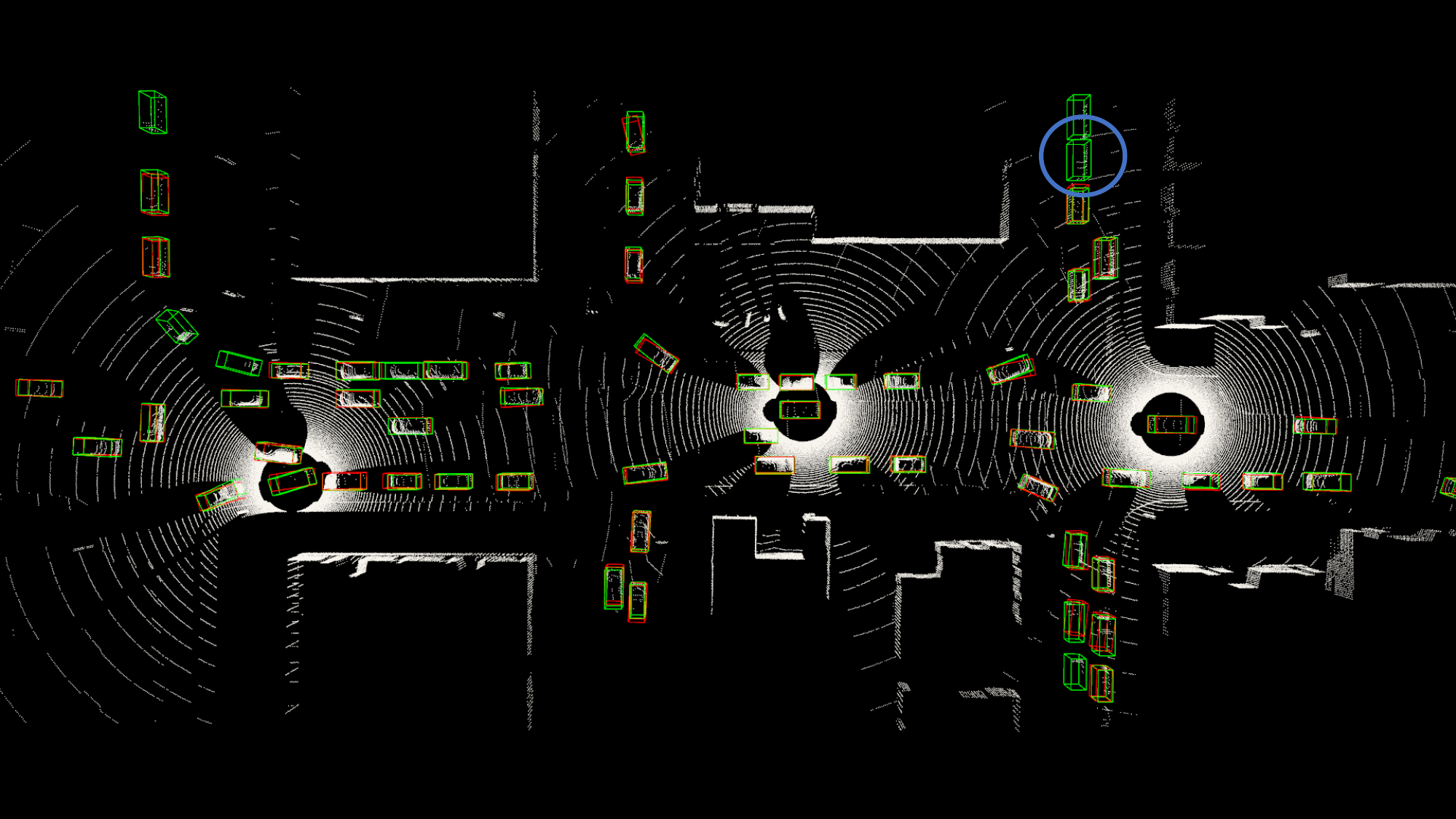}
    \label{ofdm.vis3}}
    \hfil
    \subfloat[]{\includegraphics[width=3.3in]{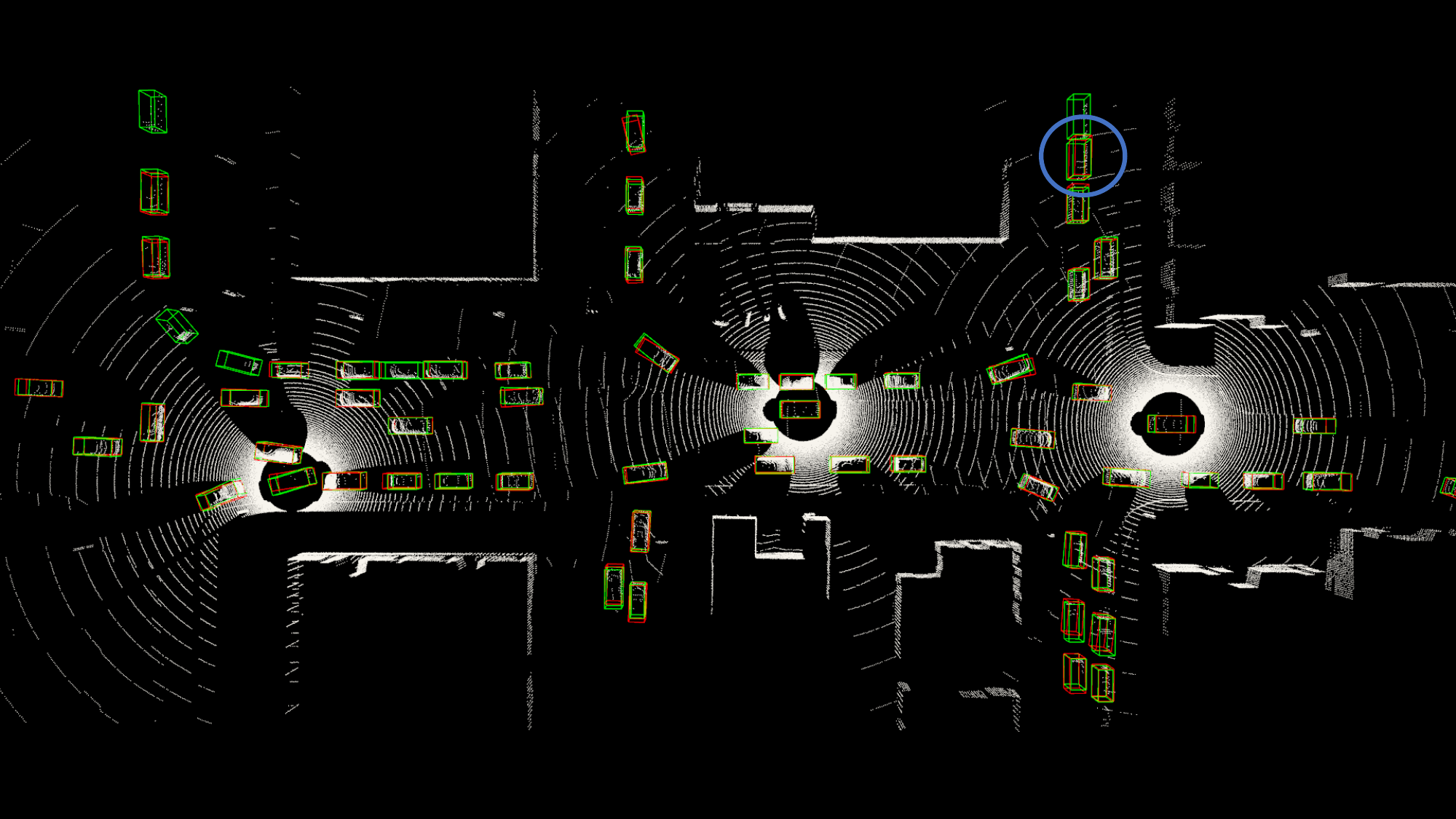}
    \label{ofdm.vis4}}
    \caption{Examples of cooperative perception with and without adaptive weighting on the WINNER II channel. The ground truth and predicted 3D bounding boxes are illustrated in green and red, respectively. Bold red circles highlight the false predictions of the cooperative perception without weighting when the SNR is 10 dB. Bold blue circles indicate the vehicle additionally detected by the weighted cooperative perception when the SNR is 30 dB.  (a) SNR = 10 dB, without weighting. (b) SNR = 10 dB, with weighting. (c) SNR = 30 dB, without weighting. (d) SNR = 30 dB, with weighting.}
    \label{fig:vis}
\end{figure*}

To evaluate the adaptive weighting in a real-world dataset and its scalability to different backbones, Fig. \ref{fig:v2v4real} demonstrates the performance of PointPillars \cite{pointpillars} on the V2V4Real dataset \cite{10203124} with WINNER II channel. V2VNet \cite{Wang2020V2VNetVC} and attentive fusion\cite{9812038} are used to account for different intermediate fusion schemes.

In Fig. \ref{fig:v2v4real}, the performance of the weighted attentive fusion with PointPillars ranges from around 47\% to 70\% for IoU=0.3 and 23\% to 34\% for IoU=0.7, which performs better than the non-weighted baselines and the single-vehicle detection. As the SNR decreases from 30 dB to -10 dB, the non-weighted collaboration experiences a significant performance degradation to be below 10\%; nevertheless, the performance of the weighted model decreases slowly until it approaches the single-vehicle detection accuracy with no fusion. This indicates that the proposed weighting algorithm could effectively work with PointPillars and adapt to different channel conditions in real-world datasets. Similar observations can be made for the performance of the weighted V2VNet with PointPillars in Fig. \ref{fig:v2v4real}. The weighted V2VNet with PointPillars outperforms the non-weighted collaboration and the single-vehicle detection by mitigating the negative effects of severe channel distortions and maintaining a close accuracy to the non-weighted collaboration when the SNR is over 20 dB. These observations demonstrate that the proposed adaptive weighting can not only generalize on the real-world dataset with communication channel distortion but is also compatible with different intermediate fusion schemes and detection backbones.

\subsection{Visualization example of adaptive weighting}\label{sec4.6}
In order to visualize how adaptive weighting improves cooperative perception, Fig. \ref{fig:vis} shows examples of cooperative perception with and without weighting. When the SNR is 10 dB, the baseline without weighting is affected by the channel distortion, thus leading to false predictions shown in Fig. \subref{ofdm.vis1}. However, it is demonstrated in Fig. \subref{ofdm.vis2} that adaptive weighting could eliminate the falsely predicted bounding boxes. When the SNR is 30 dB, both weighted and non-weighted systems can significantly reduce the false predictions due to the improved channel conditions, which are demonstrated in Fig. \subref{ofdm.vis3} and Fig. \subref{ofdm.vis4}. Nevertheless, the weighted cooperative perception detects an additional vehicle, highlighted by the blue circle in Fig. \subref{ofdm.vis4}, which the non-weighted baseline neglects. This indicates that the proposed weighting algorithm can not only mitigate the adverse effects under severe channel conditions, but also enhance the received shared features when the channel condition improves.  

\section{Conclusion} \label{sec5}
In this work, we have proposed a self-supervised adaptive weighting model for intermediate fusion to mitigate the adverse effects of distorted information caused by channel impairments. The performance of our proposed weighting algorithm has been evaluated in terms of different datasets, detection backbones, fusion schemes, noise levels and path loss factors. Numerical results and visualization examples have demonstrated that the proposed adaptive weighting algorithm performs better than the benchmarks without weighting under all conditions. It is shown that the proposed weighting algorithm can mitigate the adverse effects of severe signal distortion and enhance the received shared features when the channel condition improves. Moreover, the adaptive weighting algorithm has also demonstrated good generalization to unseen realistic communications channels and test datasets from different domains. It is also validated that the proposed weighting can flexibly scale to other detection backbones, fusion schemes and real-world datasets.  

\ifCLASSOPTIONcaptionsoff
  \newpage
\fi





\begin{thebibliography}{10}
\providecommand{\url}[1]{#1}
\csname url@samestyle\endcsname
\providecommand{\newblock}{\relax}
\providecommand{\bibinfo}[2]{#2}
\providecommand{\BIBentrySTDinterwordspacing}{\spaceskip=0pt\relax}
\providecommand{\BIBentryALTinterwordstretchfactor}{4}
\providecommand{\BIBentryALTinterwordspacing}{\spaceskip=\fontdimen2\font plus
\BIBentryALTinterwordstretchfactor\fontdimen3\font minus
  \fontdimen4\font\relax}
\providecommand{\BIBforeignlanguage}[2]{{%
\expandafter\ifx\csname l@#1\endcsname\relax
\typeout{** WARNING: IEEEtran.bst: No hyphenation pattern has been}%
\typeout{** loaded for the language `#1'. Using the pattern for}%
\typeout{** the default language instead.}%
\else
\language=\csname l@#1\endcsname
\fi
#2}}
\providecommand{\BIBdecl}{\relax}
\BIBdecl

\bibitem{9963987}
L.~Chen, \emph{et al.}, ``Milestones in
  autonomous driving and intelligent vehicles: Survey of surveys,'' \emph{IEEE Trans. Intell. Veh.}, vol.~8, no.~2, pp. 1046--1056, 2023.

\bibitem{8578570}
Y.~Zhou and O.~Tuzel, ``Voxelnet: End-to-end learning for point cloud based 3d
  object detection,'' in \emph{Proc. IEEE Conf. Comput. Vis. Pattern Recognit.}, 2018, pp. 4490--4499.


\bibitem{pointpillars}
A.~H. Lang, S.~Vora, H.~Caesar, L.~Zhou, J.~Yang, and O.~Beijbom,
  ``Pointpillars: Fast encoders for object detection from point clouds,'' in
  \emph{Proc. IEEE Conf. Comput. Vis. Pattern Recognit.}, 2019, pp. 12\,689--12\,697.

\bibitem{s18103337}
Y.~Yan, Y.~Mao, and B.~Li, ``Second: Sparsely embedded convolutional
  detection,'' \emph{Sensors}, vol.~18, no.~10, 2018.

\bibitem{8885377}
Q.~Chen, S.~Tang, Q.~Yang, and S.~Fu, ``Cooper: Cooperative perception for
  connected autonomous vehicles based on 3d point clouds,'' in \emph{Proc. IEEE Int. Conf. Distrib. Comput. Syst.}, 2019, pp. 514--524.


\bibitem{10.1145/3318216.3363300}
Q.~Chen, X.~Ma, S.~Tang, J.~Guo, Q.~Yang, and S.~Fu, ``F-cooper: Feature based
  cooperative perception for autonomous vehicle edge computing system using 3d
  point clouds,'' in \emph{Proc. IEEE/ACM Symp. Edge Comput.}, 2019, pp. 88–100.


\bibitem{Wang2020V2VNetVC}
T.-H. Wang, S.~Manivasagam, M.~Liang, B.~Yang, W.~Zeng, and R.~Urtasun,
  ``V2vnet: Vehicle-to-vehicle communication for joint perception and
  prediction,'' in \emph{Proc. Eur. Conf. Comput. Vis.}, 2020, pp. 605–621.


\bibitem{9156848}
Y.~Liu, J.~Tian, N.~Glaser, and Z.~Kira, ``When2com: Multi-agent perception via
  communication graph grouping,'' in \emph{Proc. IEEE Conf. Comput. Vis. Pattern Recognit.}, 2020, pp. 4105--4114.

\bibitem{where2comm}
Y.~Hu, S.~Fang, Z.~Lei, Y.~Zhong, and S.~Chen, ``Where2comm:
  Communication-efficient collaborative perception via spatial confidence
  maps,'' in \emph{Proc. Adv. Neural Inf. Process. Syst.}, vol.~35,
  2022, pp. 4874--4886.


\bibitem{li2021learning}
Y.~Li, S.~Ren, P.~Wu, S.~Chen, C.~Feng, and W.~Zhang, ``Learning distilled
  collaboration graph for multi-agent perception,'' in \emph{Proc. Adv. Neural Inf. Process. Syst.}, 2021,
  pp. 29\,541--29\,552.


\bibitem{9812038}
R.~Xu, H.~Xiang, X.~Xia, X.~Han, J.~Li, and J.~Ma, ``Opv2v: An open benchmark
  dataset and fusion pipeline for perception with vehicle-to-vehicle
  communication,'' in \emph{Proc. IEEE Int. Conf. Robot. Automat.}, 2022, pp. 2583--2589.

\bibitem{xu2022v2xvit}
R.~Xu, H.~Xiang, Z.~Tu, X.~Xia, M.-H. Yang, and J.~Ma, ``V2x-vit:
  Vehicle-to-everything cooperative perception with vision transformer,'' in
  \emph{Proc. Eur. Conf. Comput. Vis.}, 2022, pp. 107–124.

\bibitem{10077757}
J.~Li, R.~Xu, X.~Liu, J.~Ma, Z.~Chi, J.~Ma, and H.~Yu, ``Learning for
  vehicle-to-vehicle cooperative perception under lossy communication,''
  \emph{IEEE Trans. Intell. Veh.}, vol.~8, no.~4, pp.
  2650--2660, 2023.

\bibitem{Qiao2023WACV}
D.~Qiao and F.~Zulkernine, ``Adaptive feature fusion for cooperative perception
  using lidar point clouds,'' in \emph{Proc. IEEE Winter Conf. Appl. Comput. Vis.}, 2023, pp. 1186--1195.


\bibitem{9228884}
E.~Arnold, M.~Dianati, R.~de~Temple, and S.~Fallah, ``Cooperative perception
  for 3d object detection in driving scenarios using infrastructure sensors,''
  \emph{IEEE Trans. Intell. Transp. Syst.}, vol.~23,
  no.~3, pp. 1852--1864, 2022.


\bibitem{10184097}
C.~Liu, Y.~Chen, J.~Chen, R.~Payton, M.~Riley, and S.-H. Yang, ``Cooperative
  perception with learning-based v2v communications,'' \emph{IEEE Wirel. Commun. Lett.}, vol. 12, no. 11, pp. 1831-1835, 2023.

\bibitem{5710954}
C.~F.~Mecklenbrauker, \emph{et al.}, ``Vehicular channel characterization and its implications for wireless system design and performance,'' in \emph{Proc. of the IEEE}, vol. 99, no. 7, pp. 1189-1212, 2011.


\bibitem{5307472}
C.-x.~Wang, X.~Cheng, D.~I.~Laurenson, ``Vehicle-to-vehicle channel modeling and measurements: recent advances and future challenges,'' in \emph{IEEE Commun. Mag.}, vol. 47, no. 11, pp. 96-103, 2009.


\bibitem{Pillar-Based}
Y.~Wang, A.~Fathi, A.~Kundu, D.~A. Ross, C.~Pantofaru, T.~Funkhouser, and
  J.~Solomon, ``Pillar-based object detection for autonomous driving,'' in
  \emph{Proc. Eur. Conf. Comput. Vis.}, 2020, pp. 18--34.


\bibitem{8954080}
S.~Shi, X.~Wang, and H.~Li, ``Pointrcnn: 3d object proposal generation and
  detection from point cloud,'' in \emph{Proc. IEEE Conf. Comput. Vis. Pattern Recognit.}, 2019, pp. 770--779.


\bibitem{9157234}
S.~Shi, C.~Guo, L.~Jiang, Z.~Wang, J.~Shi, X.~Wang, and H.~Li, ``Pv-rcnn: Point-voxel feature set abstraction for 3d object detection,'' in \emph{Proc. IEEE Conf. Comput. Vis. Pattern Recognit.}, 2020, pp. 10\,526--10\,535.


\bibitem{9197364}
Y. -C.~Liu, J.~Tian, C. -Y.~Ma, N.~Glaser, C. -W.~Kuo and Z.~Kira, ``Who2com: Collaborative perception via learnable handshake communication,'' in \emph{Proc. IEEE Int. Conf. Robot. Automat.}, 2020, pp. 6876-6883.


\bibitem{10148929}
Z.~Meng, X.~Xia, R.~Xu, W.~Liu and J.~Ma, ``Hydro-3d: Hybrid object detection and tracking for cooperative perception using 3d lidar,'' \emph{IEEE Trans. Intell. Veh.}, vol. 8, no. 8, pp. 4069-4080, 2023.

\bibitem{XIA2023104120}
X.~Xia, \emph{et al.}, ``An automated driving systems data acquisition and analytics platform,''
  \emph{Transp. Res. Part C: Emerg. Technol.}, vol. 151, p. 104120, 2023.

\bibitem{winner2}
\BIBentryALTinterwordspacing
P.~Kyosti, et al ``Winner ii channel models,'' \emph{IST, Tech. Rep. IST-4-027756
  WINNER II D1.1.2 V1.2}, 2007. [Online]. Available:
  \url{https://cir.nii.ac.jp/crid/1571698600936523008}
\BIBentrySTDinterwordspacing

\bibitem{8237586}
T.-Y. Lin, P.~Goyal, R.~Girshick, K.~He, and P.~Dollár, ``Focal loss for dense
  object detection,'' in \emph{Proc. IEEE Int. Conf. Comput. Vis.}, 2017, pp. 2999--3007.


\bibitem{10045043}
R.~Xu, \emph{et al.}, ``The OpenCDA Open-Source Ecosystem for Cooperative Driving Automation Research,'' in \emph{IEEE Trans. Intell. Veh.}, vol. 8, no. 4, pp. 2698-2711, 2023.

\bibitem{10203124}
R. Xu, \emph{el al.}, ``V2V4Real: A real-world large-scale dataset for vehicle-to-vehicle cooperative perception,'' in \emph{Proc. IEEE Conf. Comput. Vis. Pattern Recognit.}, 2023, pp. 13712--13722.

\end{thebibliography}



\begin{IEEEbiographynophoto}{Chenguang Liu}
Chenguang Liu received his B.E. degree in software engineering from Dalian University of Technology, Dalian, P.R.China, in 2016 and M.S. degree in advanced computer science from The University of Manchester, U.K., in 2017. He is currently studying as a Ph.D. student at the University of Warwick, U.K. His research interests include deep learning, wireless communications and cooperative perception.
\end{IEEEbiographynophoto}
\begin{IEEEbiographynophoto}{Jianjun Chen}
Jianjun Chen is working toward the PhD degree in the Australian Artificial Intelligence Institute within the Faculty of Engineering and Information Technology, University of Technology Sydney, Australia. He received the BS degree in software engineering and MS degree in computer science from the Taiyuan University of Technology, China. His research interests include Federated Learning, Cooperative Perception and 3D Object Detection.
\end{IEEEbiographynophoto}
\begin{IEEEbiographynophoto}{Yunfei Chen}
Yunfei Chen (S'02-M'06-SM'10) received his B.E.and M.E. degrees in electronics engineering from Shanghai Jiaotong University, Shanghai, P.R.China,in 1998 and 2001, respectively. He received his Ph.D. degree from the University of Alberta in 2006. He is currently working as a Professor at the University of Durham, U.K. His research interests include wireless communications, cognitive radios, wireless relaying and energy harvesting.
\end{IEEEbiographynophoto}

\begin{IEEEbiographynophoto}{Ryan Payton}
Ryan Payton received his BSc degree in Environmental Geology in 2015 and PhD in Earth Sciences in 2022 from Royal Holloway, University of London. He is currently a Senior Research Advocate at Oracle within the Oracle for Research team, providing opportunities and technical support for researchers using Oracle Cloud Infrastructure across a range of scientific research disciplines.
\end{IEEEbiographynophoto}

\begin{IEEEbiographynophoto}{Michael Riley}
Michael Riley received his BSc degree in Information Systems in 1988 form North Staffordshire Polytechnic. He has worked in various Consulting roles within Oracle for the past 25 years. He is currently a Cloud Solution Architect supporting research projects that are utilising Oracle Cloud Infrastructure for their research projects, working collaboratively as part of the Oracle for Research program.  
\end{IEEEbiographynophoto}

\begin{IEEEbiographynophoto}{Shuang-Hua Yang}
Shuang-Hua Yang (Senior Member, IEEE) received the B.S. degree in instrument and automation and the M.S. degree in process control from the China University of Petroleum, Beijing, China, in 1983 and 1986, respectively, and the Ph.D. degree in intelligent systems from Zhejiang University, Hangzhou, China, in 1991. He is currently the director of the Shenzhen Key Laboratory of Safety and Security for Next Generation of Industrial Internet, Southern University of Science and Technology, and a Professor and the Head of the Department of Computer Science, University of Reading, U.K. His research interests include Internet of Things, industrial internet, Cyber-Physical System safety and security, Process Systems Engineering. Prof. Yang is a fellow of IET and InstMC, U.K. He is an Associate Editor of IET Cyber-Physical Systems: Theory and Applications.
\end{IEEEbiographynophoto}

\end{document}